\documentclass{article}

\usepackage{microtype}
\usepackage{graphicx}
\usepackage{subfigure}
\usepackage{booktabs} 
\usepackage{amsmath}
\usepackage{amssymb}
\usepackage{todonotes}

\let\mc\multicolumn

\usepackage{hyperref}

\usepackage[accepted]{icml2020}

\icmltitlerunning{Message Passing Query Embedding}

\begin{document}

\twocolumn[
\icmltitle{Message Passing Query Embedding}



\icmlsetsymbol{equal}{*}

\begin{icmlauthorlist}
\icmlauthor{Daniel Daza}{vu,uva,dl}
\icmlauthor{Michael Cochez}{vu,dl}
\end{icmlauthorlist}

\icmlaffiliation{vu}{Vrije Universiteit Amsterdam}
\icmlaffiliation{uva}{University of Amsterdam}
\icmlaffiliation{dl}{Discovery Lab, Elsevier, The Netherlands}

\icmlcorrespondingauthor{Daniel Daza}{d.dazacruz@vu.nl}

\icmlkeywords{knowledge graphs,machine learning,representation learning,query answering,graph convolutional networks}

\vskip 0.3in
]



\printAffiliationsAndNotice{}  

\begin{abstract}
Recent works on representation learning for Knowledge Graphs have moved beyond the problem of link prediction, to answering queries of an arbitrary structure. Existing methods are based on ad-hoc mechanisms that require training with a diverse set of query structures. We propose a more general architecture that employs a graph neural network to encode a graph representation of the query, where nodes correspond to entities and variables. The generality of our method allows it to encode a more diverse set of query types in comparison to previous work. Our method shows competitive performance against previous models for complex queries, and in contrast with these models, it can answer complex queries when trained for link prediction only. We show that the model learns entity embeddings that capture the notion of entity type without explicit supervision.
\end{abstract}

\section{Introduction}

Knowledge Graphs (KG) are useful data structures for encoding information from different domains, by representing entities and relations of different types between them.
Tasks of interest that can be addressed with knowledge graphs include
information retrieval \cite{dalton2014entity},
question answering \cite{vakulenko2019message,huang2019knowledge},
and natural language processing \cite{logan2019baracks}.
A common way to answer a question using a KG is to pose it as a structured query (for example, using the SPARQL query language \cite{harris2013sparql}). 
The query is then answered via logical inference, using the information present in the graph. 
However, knowledge graphs are usually incomplete, either due to the construction process, or their dynamic nature.
This means that there will be cases where these systems return \textit{no answer} for a query.

To address this problem, we follow recent works that propose to map the query and all entities in the KG to an embedding space \cite{hamilton2018gqe,wang2018towards,mai2019contextual}, where we can compute similarity scores to produce a ranked list of answers. We propose Message Passing Query Embedding (MPQE), motivated by the observation that queries over a KG can be represented by small graphs, where nodes correspond to constant entities and variables. We employ a Graph Neural Network (GNN) to perform message passing on the query graph, and an aggregation function to combine all the messages in a single vector, which acts as a representation of the query in the embedding space. 
By training on the task of query answering, our method learns jointly an embedding for each entity in the KG, and type embeddings for variables.


\begin{figure}[t]
\centering
\includegraphics[width=0.99\linewidth]{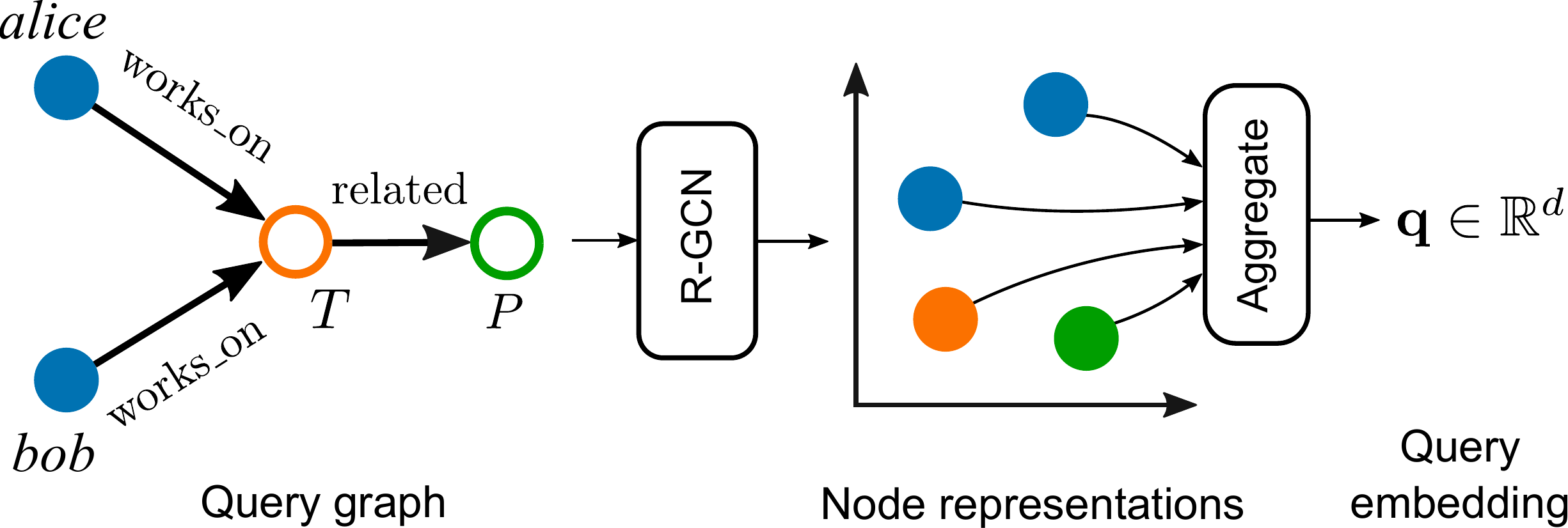}
\caption{Message Passing Query Embedding takes as input a query graph and outputs a query embedding. Features for each node in the query graph are embeddings of entities in the KG, or type embeddings. A GNN propagates information across the graph, and an aggregation function yields the query embedding.}
\label{fig:model}
\end{figure}

Our contributions can be summarized as follows: 1) We propose a novel method to embed queries over knowledge graphs, that addresses limitations of previous works in terms of computational complexity, and the diversity of query structures that it admits; 2) we show experimentally that MPQE is competitive with previous work on complex query embedding across multiple query structures. Furthermore, we provide evidence of the superior generalization of MPQE by training for link prediction only, and testing on complex queries. While previous models fail in this setting, we show that MPQE retains predictive performance on complex queries; 3) we conduct a qualitative analysis of the entity embeddings produced by query embedding methods, and show that MPQE learns in an unsupervised way a structured embedding space where entities cluster according to type.

\section{Problem Definition}

We define a Knowledge Graph (KG) as a tuple $(\mathcal{V}, \mathcal{E}, \mathcal{R}, \mathcal{T})$, where $\mathcal{V}$ is a set of nodes representing entities, and $\mathcal{E}$ a set of typed edges between the nodes.
A function  $\tau: \mathcal{V} \rightarrow \mathcal{T}$ assigns a type to every node, where $\mathcal{T}$ is a set of entity types.
Each edge in $\mathcal{E}$ corresponds to a relation between two nodes $v_i$ and $v_j \in \mathcal{V}$, that we denote by $r(v_i, v_j)$, where $r\in\mathcal{R}$ is a relation type.



Given a KG, we can pose queries that seek for an entity satisfying certain conditions.
We consider queries in \textit{conjunctive form}, formed by a conjunction of binary predicates where the arguments are entities or variables. To illustrate this, consider a KG of an academic institution where researchers work on topics, and topics are related to projects, and the following query: ``select all projects $P$, such that topic $T$ is related to $P$, and both \textit{alice} and \textit{bob} work on $T$.''
This query asks for entities $P$ that satisfy the following condition:
\begin{align*}
P.\exists T, P:~
&\mathrm{related}(P, T) \notag\\
&\wedge \mathrm{works\_on}(\text{\textit{alice}}, T) \wedge \mathrm{works\_on}(\text{\textit{bob}}, T).
\end{align*}
In general, a query $q$ is defined by a condition on a \textit{target variable} $V_t$ as follows:
\begin{equation}
q = V_t.\exists V_1, \dots, V_m: r_1(a_1, b_1) \wedge \ldots \wedge r_m(a_m, b_m),
\label{eq:query_def}
\end{equation}
where $r_i\in\mathcal{R}$, and $a_i$ and $b_i$ are either entities in the KG, or query variables in $\lbrace V_t, V_1, \dots, V_m \rbrace$. An entity is a correct answer if it satisfies the condition defined by the query. We can address the problem of returning a list of entities that satisfy the query, by mapping the query to an embedding and computing its similarity with the embeddings of entities in the KG~\citep{hamilton2018gqe}. Formally, we optimize an embedding $\mathbf{e}_v\in\mathbb{R}^d$ for every entity $v\in\mathcal{V}$, and we define an embedding method for the query that maps it to a vector $\mathbf{q}\in\mathbb{R}^d$. We then compute the cosine similarity between $\mathbf{q}$ and an entity embedding $\mathbf{e}_v$ as follows:
\begin{equation}
\mathrm{score}(\mathbf{q}, \mathbf{e}_v) = \frac{\mathbf{q}^\top\mathbf{e}_v}{\Vert \mathbf{q} \Vert \Vert \mathbf{e}_v \Vert}.
\label{eq:score}
\end{equation}
This score determines the rank of an entity as a possible answer to the query.





\section{Related Work}

Multiple approaches for machine learning on graphs consider embedding the graph into a vector space via link prediction~\cite{bordes2013translating,wang2014TransH,distmult}. 
The applicability of these methods for answering complex queries is limited. For each link that needs to be predicted, these methods must consider all possible entities, which is exponential in the size of the query. Our method is based on an architecture that directly encodes the query into an embedding, which provides our method with a linear complexity in the size of the graph.
Recent works have also addressed the problem of embedding a query to retrieve approximate answers, by partitioning the query graph in different subgraphs, so that candidate answers can be provided for each of them \cite{Zhang2018TrQueryAE}. In \cite{wang2018towards}, the authors pre-train embeddings using an algorithm inspired by TransE. Instead of relying on a separate pre-training step, we learn with an objective that optimizes entity embeddings for the task of query answering directly.
The most related approaches to our work consider embedding queries directly in the embedding space~\cite{hamilton2018gqe,mai2019contextual,ren2020query2box}, by applying a sequence of projection and intersection operators that follow the structure of the query graph. These methods are constrained to having entities only at the leaves of the Directed Acyclic Graph (DAG) defined by the query. Furthermore, the use of projection and intersection mechanisms requires training with multiple query structures that comprise both chains and intersections. Our method has a more general formulation that enables it to i) encode a general set of query graphs, without constraints on the location of entities in the query, and ii) learn from single-hop link prediction training alone, and still generalize to larger queries.

\section{Message Passing Query Embedding}

As noted in previous work \cite{hamilton2018gqe,wang2018towards}, queries in conjunctive form can be represented as a DAG. 
In this graph, the leaf nodes correspond to constant entities, the root to the variable to be retrieved, and any intermediate nodes to other variables in the query. Given a query of the form given in eq. \ref{eq:query_def}, we define the \textit{query graph} as the tuple $(\mathcal{V}_q, \mathcal{E}_q, \mathcal{R}_q)$. 
Here, $\mathcal{V}_q$ is the union of nodes for constant entities $\mathcal{V}_{qe}$ and variables $\mathcal{V}_{qv}$ in the query. To construct the set of edges $\mathcal{E}_q$, we add one edge $r_i(a,b)$ for each predicate in the query.

\subsection{Model Definition}

Message Passing Query Embedding (MPQE) learns a matrix of \textit{entity embeddings} $\mathbf{M}_e\in\mathbb{R}^{\vert \mathcal{V}\vert\times d}$, where each row contains an embedding for each entity in the KG, and $d$ is the dimension of the embedding space. Additionally, it also learns a matrix of \textit{type embeddings} $\mathbf{M}_t\in\mathbb{R}^{\vert\mathcal{T}\vert\times d}$, with one embedding for each type of entity in the KG. Assume we define an ordering on the sets $\mathcal{E}$ and $\mathcal{T}$. The function $\phi_e(v)$ returns the row of $\mathbf{M}_e$ corresponding to $v$, for all entities $v\in\mathcal{E}$. Similarly, for all types $v\in\mathcal{T}$, $\phi_t(v)$ returns its corresponding row of $\mathbf{M}_t$.

Given a query graph, we start by initializing each of the nodes with a vector representation $\mathbf{h}_v^{(0)}$, from $\mathbf{M}_e$ if it corresponds to a constant entity, or from $\mathbf{M}_t$ for a variable. Formally, $\mathbf{h}_v^{(0)} = \phi_e(v)$ if $v\in\mathcal{V}_{qe}$, and $\mathbf{h}_v^{(0)} = \phi_t(v)$ if $v\in\mathcal{V}_{qv}$. We proceed by applying $L$ steps of \emph{message passing} with a a Relational Graph Convolutional Network (R-GCN) \cite{schlichtkrull2018rgcn}, which updates the representation of a node taking into account its neighbors and the type of the relations involved.
The representation for node $v$ at step $l+1$ in the R-GCN is defined as follows:
\begin{equation}
\mathbf{h}^{(l+1)}_v = f\left( \mathbf{W}^{(l)}_0\mathbf{h}^{(l)}_v + \sum_{r\in\mathcal{R}} \sum_{j\in\mathcal{N}_v^r}\frac{1}{\vert \mathcal{N}_v^r\vert}\mathbf{W}^{(l)}_r \mathbf{h}^{(l)}_j  \right),
\label{eq:rgcn}
\end{equation}
where $f$ is a non-linearity, $\mathcal{N}_v^r$ is the set of neighbors of node $v$ through relation type $r$, and $\mathbf{W}_0^{(l)}$ and $\mathbf{W}_r^{(l)}$ are parameters of the model. After $L$ applications of an R-GCN layer, the node representations are combined into a single vector that acts as the embedding of the query, by means of an \emph{aggregation} function $\phi$:
\begin{equation}
\mathbf{q}_\phi = \phi\left(\lbrace \mathbf{h}^{(L)}_v \vert v \in \mathcal{V}_{qe} \cup\mathcal{V}_{qv} \rbrace\right).
\end{equation}

\paragraph{Fixed message passing} We consider as aggregation functions sum and max pooling over all $\mathbf{h}_v^{(L)}$. We also experiment with the concatenation of representations from hidden layers of the R-GCN, followed by an MLP and a sum over nodes in the query graph. We denote this function as CMLP:
\begin{equation}
\mathbf{q}_\text{CMLP} = \sum_{v\in\mathcal{V}_q}\text{MLP}\left([\mathbf{h}^{(1)}_v, \dots, \mathbf{h}^{(L)}_v]\right).
\end{equation}

\paragraph{Dynamic embedding} Let $D$ denote the \textit{diameter} of the query graph (the longest shortest path between two nodes in the graph). We propose a dynamic query embedding method, by noting that at most $D$ message passing steps are required to propagate messages from all nodes, to the target node. The method performs $D$ steps of message passing, and it then selects the representation $\mathbf{h}_{v_t}^{(D)}$ of the target node $v_t$ as the embedding of the query. We denote this as the Target Message (TM) function.

\textbf{Training} Following previous work on query embedding \cite{hamilton2018gqe}, we optimize all parameters using gradient descent on a contrastive loss function, where given a query $q$ and its embedding $\mathbf{q}$, a positive sample $v^+$ corresponds to an entity in the knowledge graph that answers the query, and a negative sample $v^-$ is an entity sampled at random, that is not an answer to the query but has the correct type.
We minimize a margin loss function:
\begin{equation}
\mathcal{L}(q) = \max(0, 1 - \text{score}(\mathbf{q},\mathbf{h}_{v^+}^{(0)}) + \text{score}(\mathbf{q},\mathbf{h}_{v^-}^{(0)})).
\label{eq:loss}
\end{equation}

\section{Experiments}

We evaluate the performance of MPQE in query answering over knowledge graphs, by considering 7 different query structures (detailed in Appendix \ref{app:structures}).
All the code to reproduce our experiments is available online
\footnote{\url{https://github.com/dfdazac/mpqe}}.

\paragraph{Datasets} We use publicly available knowledge graphs that have been used in the literature of graph representation learning and query answering \cite{ristoski2016collection,hamilton2018gqe} containing from thousands to millions of entities -- \textbf{AIFB}: a KG of an academic institution, where entities are persons, organizations, projects, publications, and topics; \textbf{MUTAG}: a KG of carcinogenic molecules, where entities are atoms, bonds, compounds, and structures; \textbf{AM}: contains relations between different artifacts in the Amsterdam Museum, including locations, makers, and others; \textbf{Bio}: a dataset of a biological interaction network containing entities of type drug, disease, protein, side effect, and biological processes. Their statistics can be found in Appendix \ref{app:stats}.

\paragraph{Query generation} We follow the evaluation procedure of \citet{hamilton2018gqe}: to obtain query graphs, we sample subgraphs from the KG. Each subgraph specifies the entities and the types of variables in the query, and the correct answer to the query. For each query we also obtain a negative sample, and in the case of query graphs with intersections, a \textit{hard} negative sample. This is an entity that would be a correct answer, if the conjunction represented by the intersection is relaxed to a disjunction. Before sampling subgraphs for training, we remove edges from the KG. We then guarantee that subgraphs sampled to generate query graphs for testing rely on at least one of these remove edges.

\paragraph{Experimental setup} With the exception of the TM aggregation function (where the number of message passing steps is given by the query diameter), we use 2 R-GCN layers. For aggregation functions with MLPs, we use two fully-connected layers, and in all cases we use ReLU for the nonlinearities. We minimize eq. \ref{eq:loss} using the Adam optimizer with a learning rate of 0.01, and use an embedding dimension of 128. We train the models for 1-hop link prediction until convergence, and then on the full set of query structures. As a baseline we evaluate the Graph Query Embedding (GQE) method by \citet{hamilton2018gqe} with their TransE, DistMult, and Bilinear variants.

\subsection{Results}

\paragraph{Query answering} We report the area under the ROC curve (AUC) and the Average Percentile Rank (APR) on the test set. The results for the query answering task are shown in Table \ref{tab:results-all}. We observe that MPQE obtains competitive performance in comparison with GQE across different datasets. MPQE underperforms in the MUTAG dataset, which we identified as the dataset with the least diverse set of relations. We noticed that while GQE-DistMult handles hard negative samples well (which occur only on queries with intersections), MPQE-TM has better performance on regular samples, across all query structures. We discuss further interesting properties of MPQE in Appendix \ref{sec:properties}.

\begin{table}[t]
\caption{Results on query answering averaged across different query structures. Highlighted values denote the best variant within each group of GQE and MPQE models.}
\label{tab:results-all}
\vskip 0.15in
\centering
\resizebox{0.48\textwidth}{!}{
\begin{tabular}{lcccccccc}
\midrule
             & \mc{2}{c}{AIFB}   &  \mc{2}{c}{MUTAG}  &  \mc{2}{c}{AM}     & \mc{2}{c}{Bio}        \\ 
               \cmidrule(lr){2-3}  \cmidrule(lr){4-5}  \cmidrule(lr){6-7}    \cmidrule(lr){8-9}
Method       &     AUC &     APR &     AUC &    APR   &    AUC &     APR   &     AUC &     APR     \\ 
\midrule
GQE-TransE   &    83.1 &\bf 86.7 &    78.8 &    81.0  &    80.9 &    82.3  &    87.4 &    88.9     \\ 
GQE-DistMult &\bf 83.8 &    86.0 &\bf 80.6 &\bf 81.1  &\bf 82.9 &    83.2  &    90.0 &    90.3     \\ 
GQE-Bilinear &    83.4 &    83.3 &    78.5 &    79.7  &    80.7 &\bf 84.4  &\bf 90.5 &\bf 90.8     \\ 
\midrule
MPQE-TM      &    84.9 &    87.4 &\bf 76.7 &\bf 77.6  &\bf 84.2 &\bf 86.3  &    88.8 &    89.8     \\ 
MPQE-sum     &    84.7 &    86.8 &    74.6 &    73.1  &    80.9 &    83.6  &    90.0 &\bf 90.5     \\ 
MPQE-max     &    83.4 &    85.9 &    74.9 &    72.7  &    80.9 &    82.5  &    88.3 &    88.7     \\
MPQE-CMLP    &\bf 86.3 &\bf 89.1 &    74.3 &    72.5  &    82.5 &    85.5  &\bf 90.1 &    90.2     \\
\midrule
\end{tabular}
}
\end{table}

\paragraph{Generalization} To examine the generalization properties of MPQE, we train the models on simple queries that require 1-hop link prediction, but we carry out the evaluation using the complete set of complex query structures. Unlike MPQE, in this case GQE cannot provide an answer better than random for queries with intersections, because the intersection operator is not optimized. We thus consider two evaluations modes: evaluating on queries with chain structures only, and evaluating on the complete set of query structures (where GQE is not applicable). 
These modes are denoted as ``ch'' and ``all'', respectively, in Table \ref{tab:generalization}. 
The results of MPQE are competitive when evaluating on queries with chains only, and crucially, it also generalizes well to query structures not seen during training.
This encouraging results shows that MPQE implements a mechanism that does not necessarily require training on many diverse query structures to generalize well, unlike GQE.

\begin{table}[t]
\caption{Generalization results on complex query answering (AUC) when training for simple link prediction. The results show the performance on a test set of queries with chains only (ch), and the complete set of queries with chains and intersections (all). Dashes indicate not better than random results.}
\label{tab:generalization}
\vskip 0.15in
\centering
\resizebox{0.48\textwidth}{!}{
\begin{tabular}{lcccccccc}
\toprule
             &   \multicolumn{2}{c}{AIFB} &   \multicolumn{2}{c}{MUTAG} &   \multicolumn{2}{c}{AM} &  \multicolumn{2}{c}{Bio} \\
Method       &   ch        &     all      &   ch         &         all  &   ch     &       all     &   ch     &       all     \\ 
\midrule                                                                                                             
GQE-TransE   &    74.0     &     ---      &\bf 89.4      &         ---  &    85.8  &         ---   &    85.5  &         ---   \\ 
GQE-DistMult &    72.8     &     ---      &    85.4      &         ---  &    82.4  &         ---   &    95.9  &         ---   \\ 
GQE-Bilinear &    72.7     &     ---      &    89.1      &         ---  &\bf 85.9  &         ---   &    85.8  &         ---   \\ 
\midrule                                                                                                                
MPQE-TM      &\bf 77.0     &\bf 75.5      &    86.8      &    \bf 77.2  &\bf 85.0  &    \bf 81.6   &\bf 96.4  &    \bf 83.9   \\ 
MPQE-sum     &    69.8     &    69.6      &    82.8      &        74.0  &    52.5  &        53.9   &    92.4  &        80.0   \\ 
MPQE-max     &    74.1     &    71.9      &    77.1      &        71.6  &    51.2  &        53.0   &    92.0  &        79.9   \\
MPQE-CMLP    &    69.7     &    69.1      &    84.6      &        74.2  &    51.5  &        53.8   &    89.8  &        78.3   \\ 
\bottomrule
\end{tabular}
}
\end{table} 

\begin{figure}[t]
\centering
\includegraphics[width=0.99\linewidth]{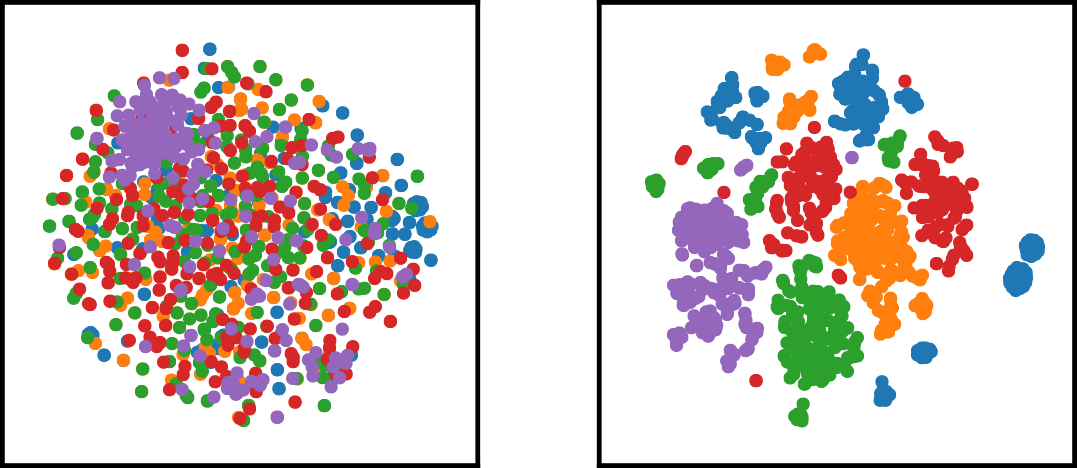}
\caption{Visualization of the entity embeddings learned by GQE (left) and MPQE (right).
Each color represents an entity type.}
\label{fig:embeddings}
\end{figure}

\paragraph{Clustering} We sample 200 entity embeddings for each type in the AM dataset, and visualize them using T-SNE~\cite{maaten2008visualizing} in Fig.~\ref{fig:embeddings} for GQE-Bilinear and MPQE-TM. We observe that MPQE has learned a structured space where entities cluster according to their type, without explicit supervision. This is in stark contrast with the embeddings of GQE, where we do not observe such a clear structure.
We hypothesize that the structured embedding space of MPQE contributes to the generalization from simple link prediction to complex queries.

\section{Conclusion}

We have presented MPQE, a neural architecture to encode complex queries on knowledge graphs, that jointly learns entity and type embeddings and a straightforward message passing architecture to obtain a query embedding. Our experiments show that message passing across the query graph is a powerful mechanism for query answering, that generalizes to multiple query structures even when only trained for single hop link prediction. Qualitative results show that MPQE learns a well-structured embedding space. This result motivates future research on the application of the learned embeddings to other tasks related to KGs, such as clustering, and node and graph classification. Under this new light, MPQE can be seen as a new method for unsupervised representation learning in KGs.

While our generalization experiments highlight the general formulation of our method, we further plan to evaluate its performance on queries not restricted to constants at the leaves of the query graph in future work. By being able to encode queries independent of the position of entities and variables, we could encode queries with additional information, that could be used to condition the answers on a given context. Such an application would be useful in information retrieval and recommender systems.

Our method presents limitations when evaluating on hard negative samples.
Further improvements could include improving the message passing procedure via adding attention or gating mechanisms,
and extensions to more expressive query embedding representations, such as boxes \citep{ren2020query2box}.


\section*{Acknowledgments}

This project was funded by Elsevier's Discovery Lab. We thank Frank van Harmelen and Erman Acar for helpful discussions and comments.

\bibliography{mpqa-icml-grl}
\bibliographystyle{icml2020}

\clearpage
\appendix

\section{Query structures}
\label{app:structures}

\begin{figure}[t]
\centering
\includegraphics[width=0.7\linewidth]{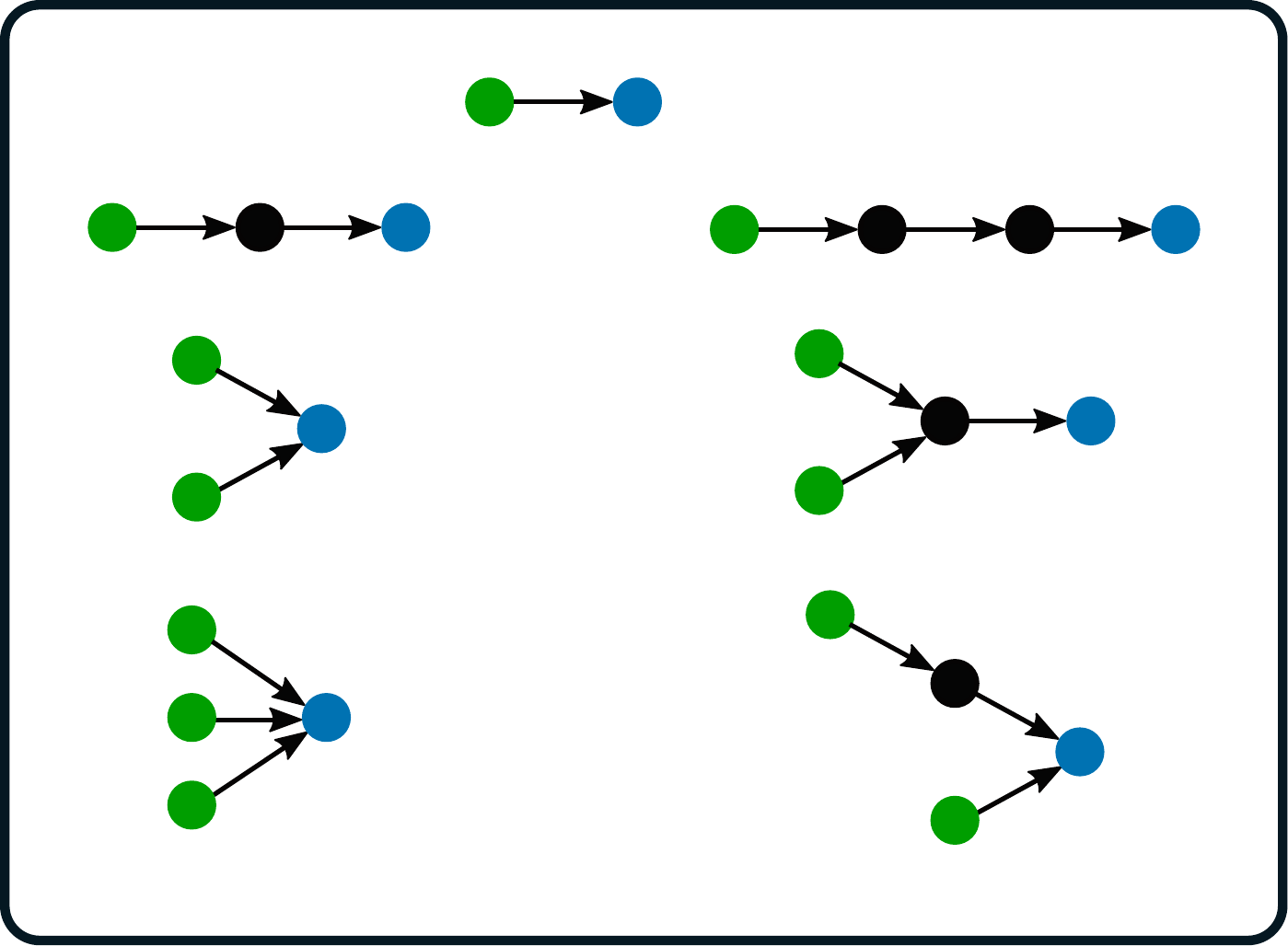}
\caption{Query structures that we consider for the evaluation of methods on query answering. Green nodes correspond to entities, and the rest are variables in the query, with blue nodes representing the target of the query.}
\label{fig:structures}
\end{figure}

We run experiments over 7 different query structures that combine chains and intersections, as detailed in Fig. \ref{fig:structures}.

\section{Dataset statistics}
\label{app:stats}

\begin{table}[t]
\caption{Statistics of the knowledge graphs that we use for training and evaluation.}
\label{tab:stats}
\vskip 0.15in
    \centering
    \resizebox{0.49\textwidth}{!}{
    \begin{tabular}{lrrrr}
    \toprule
                   & AIFB    & MUTAG  & AM        & Bio       \\
    \midrule
    Entities       &  2,601  & 22,372 &   372,584 &   162,622 \\
    Entity types   &      6  &      4 &         5 &         5 \\
    Relations      & 39,436  & 81,332 & 1,193,402 & 8,045,726 \\
    Relation types &     49  &      8 &        19 &        56 \\
    \bottomrule
    \end{tabular}
    }
\end{table}

We present statistics for the datasets used in our work in Table \ref{tab:stats}. To generate splits for training, test, and validation, we follow the procedure of \citet{hamilton2018gqe}. Given a KG, we start by removing 10\% of its edges. Using this incomplete graph, we extract 1 million subgraphs, containing all the query structures outlined previously. We then restore the removed edges, and extract 11,000 additional subgraphs of all structures, ensuring that they all rely in at least one of the edges that was removed to create the training set. We split this set of query graphs into two disjoint sets, containing 1,000 queries for validation, and 10,000 for testing. We use the validation set to perform early stopping during training.

\section{MPQE properties}
\label{sec:properties}

\begin{figure*}[t]
\centering
\includegraphics[width=0.98\linewidth]{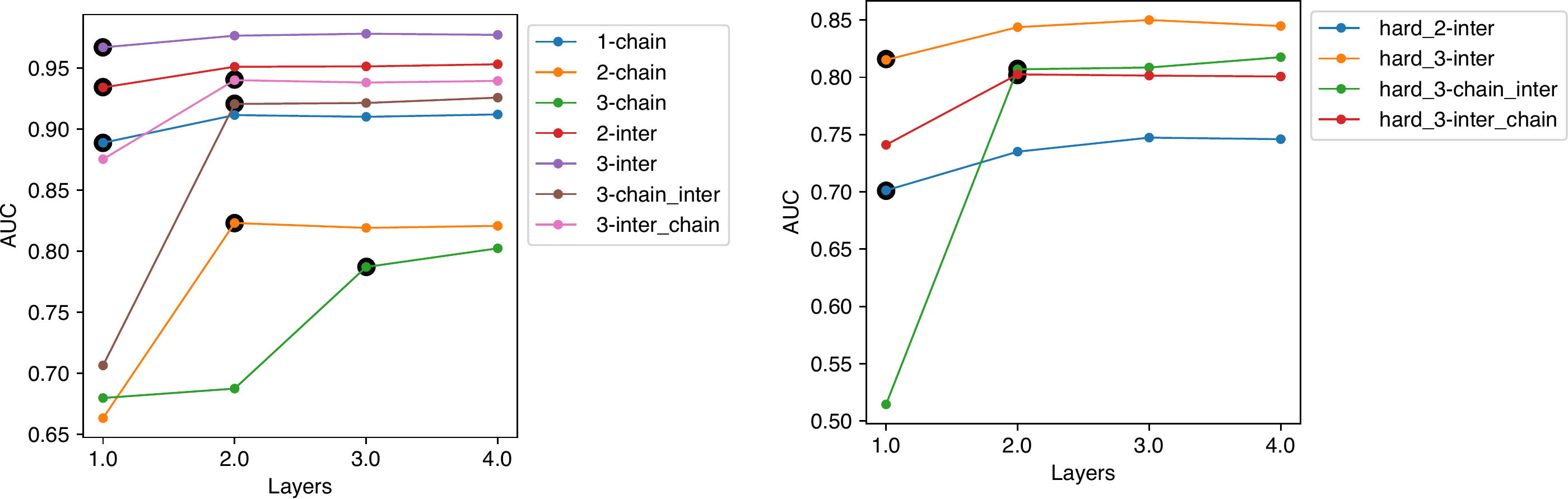}
\caption{Query answering performance (AUC) as a function of the number of message passing steps (implemented by layers of an R-GCN), evaluated across different query types. Dark circles corresponds to the diameter of the corresponding query. When the number of steps matches the diameter, there is a significant increase in performance, and further steps have little effect.}
\label{fig:mplayers}
\end{figure*}

An interesting observation from our experiments is that the message passing mechanism alone is sufficient to provide good performance for query answering, as we can see from the results for the MPQE-TM architecture. In this model, we perform a number of steps of message passing equal to the diameter of the query, and take as query embedding the resulting feature vector at the target node.
Intuitively, this allows MPQE-TM to adapt to the structure of a query so that after message passing, all information from the entity and variable nodes has reached the target node.
To confirm this intuition, we evaluate the performance of MPQE as a function of the number of message passing steps, ranging from 1 to 4. The results are shown in Figure \ref{fig:mplayers}, for all the query structures that we have considered. We highlight the points that correspond to the diameter of the query, and we note that the results align with our intuition about the message passing mechanism. When the number of steps matches the diameter, there is a significant increase in performance, and further steps have little effect. This supports the superior generalization observed in our experiments, in comparison with GQE, and other MPQE architectures where the number of R-GCN layers was fixed.

\end{document}